\documentclass{article}

\usepackage[final]{ewrl_2026}

\usepackage[utf8]{inputenc} 
\usepackage[T1]{fontenc}    
\usepackage[colorlinks=true,allcolors=blue,final]{hyperref}       
\usepackage{url}            
\usepackage{booktabs}       
\usepackage{amsfonts}       
\usepackage{nicefrac}       
\usepackage{microtype}      
\usepackage{xcolor}         
\usepackage{graphicx}       
\usepackage{subcaption}     
\usepackage{dirtree}        
\usepackage{makecell}

\usepackage{paralist}

\newcommand{\fakepar}[1]{\vspace{1mm}\noindent\textbf{#1}\quad}

\title{RLLBC-Lib: An Educational Code Library for Reinforcement Learning and Learning-Based Control}

\author{%
  Bernd Frauenknecht, \quad Emma Cramer, \quad  Artur Eisele,  \quad Paul Kruse,\\
  \textbf{Lukas Kesper, \quad Jonas Hertrampf, \quad Ramil Sabirov, \quad Jyotirmaya Patra,}\\
   \textbf{Johannes Berger}, \quad \textbf{Paul Brunzema, \quad Friedrich Solowjow, \quad Sebastian Trimpe}\\[1ex]
  Institute for Data Science in Mechanical Engineering (DSME)\\
  RWTH Aachen University\\
  52062 Aachen, Germany \\[1ex]
  \texttt{\{firstname\}.\{lastname\}@dsme.rwth-aachen.de} \\
}

\makeatletter
\renewcommand{\@notice}{}
\newcommand{\firstpagefooter}[1]{%
  \renewcommand{\@notice}{%
    \enlargethispage{2\baselineskip}%
    \@float{noticebox}[b]%
      \footnotesize #1%
    \end@float%
  }%
}
\makeatother

\firstpagefooter{Presented at the 19th European Workshop on Reinforcement Learning (EWRL 2026) }

\begin{document}

\maketitle

\begin{abstract}
    Reinforcement learning (RL) is an exciting concept as well as a remarkable success story worth sharing. However, RL builds on rather complex interactions between different objects that play out over several cycles. Such dynamics are often best explained with an easily accessible implementation. We present RLLBC-Lib, a carefully crafted code library with the goal of lowering the entry barrier for students and other learners of RL in the context of learning-based control. At its heart, RLLBC-Lib comprises a comprehensive library of tabular RL approaches to enforce a clear understanding of the theoretical foundations. A deep RL library follows the same design principles, underscoring the parallels between simple tabular and state-of-the-art deep RL approaches. Additionally, RLLBC-Lib provides a collection of implementations illustrating core RL principles and contrasting RL to other learning-based control approaches. Finally, RLLBC-Lib provides an ideal basis for creating programming assignments with automated grading.
\end{abstract}
\section{Introduction}
Deep Reinforcement Learning (RL) has achieved remarkable success, yielding state-of-the-art performance in problems such as locomotion~\citep{pmlr-v164-rudin22a}, drone racing~\citep{kaufmann_champion-level_2023}, or nuclear fusion~\citep{degrave_magnetic_2022}. Further, RL remains highly relevant in fields like fine-tuning of large language models \cite{ouyang2022training}. This relevance, together with its conceptual elegance, makes RL a compelling subject for modern university curricula.

However, we observe a substantial entry barrier for students to the field. RL algorithms are challenging to grasp from equations alone, because their behavior emerges from iterative feedback between data collection, value estimation, and policy improvement.
Hands-on implementations expose these feedback loops directly. Students can trace how experience is generated, how updates change value estimates or policies, and how choices such as exploration, bootstrapping, and function approximation affect learning. Such access is crucial for understanding both tabular methods and modern deep RL approaches.
Great deep RL libraries are available, e.g., \cite{liang2018rllib, raffin2021stable, d2021mushroomrl, huang2022cleanrl}. In contrast, educational libraries that span the full arc from tabular methods to state-of-the-art deep RL are considerably scarce. We, however, consider a solid understanding of the theoretical foundations established in the former critical to build intuition for the latter.

Furthermore, the sequential decision-making problems addressed by RL typically fall within the broader scope of learning-based control (LBC). Broadly speaking, LBC denotes the use of machine learning in the design or operation of control systems \cite{schoellig2026decade}. While RL offers a highly data-driven, machine learning-centric approach to these challenges, a variety of control-theoretic methods address similar issues.  Presenting RL in the broader context of LBC methods and discussing the respective strengths and weaknesses of different approaches builds intuition. Students learn when a problem is well suited for RL and when it is more sensible to resort to another method.

A particularly compelling aspect of teaching RL is that its applications fall into the intersection of computer science and engineering. In practice, we observe that these two cohorts have distinct pedagogical needs. Engineering students benefit immensely from easily accessible algorithmic implementations, whereas computer science students derive significant value from the control perspective, which is often not taught in standard computer science curricula.

Combining these observations, we present RLLBC-Lib, an educational code library providing
\begin{compactitem}
\item implementations of common dynamic programming and tabular RL algorithms;
\item implementations of common deep RL algorithms; accompanied by 
\item additional explanatory examples following the didactic concept of \cite{sutton2018reinforcement};
\item implementations of several orthogonal LBC approaches; and
\item a suitable starting point to create programming assignments with automated grading. 
\end{compactitem}

\begin{figure}[tb]
\vspace{3mm}
  \centering
    \includegraphics[width=\linewidth]{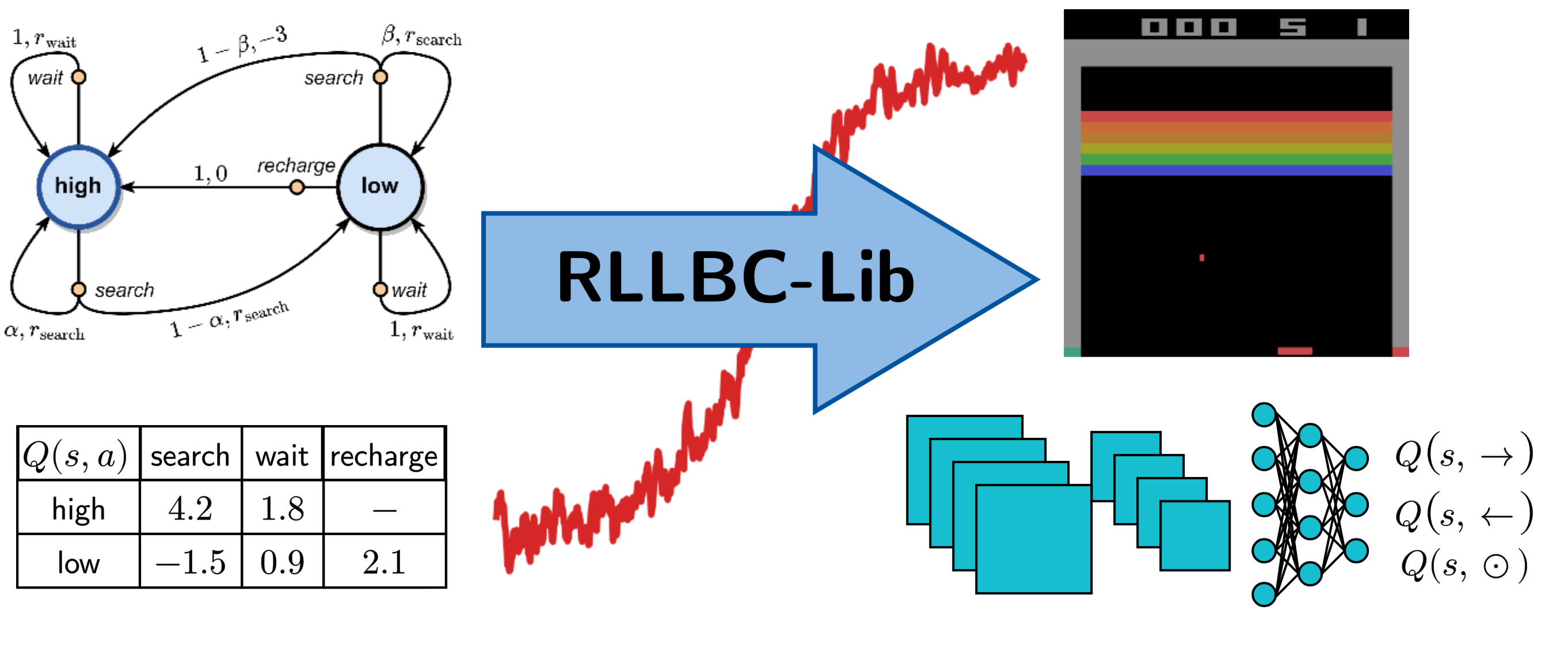}
  \caption{Didactic Concept of RLLBC-Lib. \textit{The presented code library lowers the entry barrier for students and other learners of RL. It starts from straightforward tabular settings, depicted on the left, to build a solid understanding of theoretical foundations. From there, it gradually guides towards state-of-the-art deep RL approaches, depicted on the right. This learning process is represented by the stylized return curve in red. The library builds on Jupyter notebooks with extensive explanations of theoretical concepts and their implementation. The notebook structure is consistent between tabular and deep RL approaches to underscore their similarities.}}
  \label{fig:fig1}
\end{figure}

As depicted in Figure \ref{fig:fig1}, RLLBC-Lib introduces students to the world of RL starting from simple toy examples that build intuition for basic concepts. From there, it gradually guides them towards state-of-the-art deep RL methods through accessible code accompanied by detailed explanations. In our own experience from classroom and online teaching, it is a valuable addition to RL textbooks \cite{sutton2018reinforcement, Murphy2024Dec} and the current landscape of RL teaching libraries summarized in Appendix \ref{app:lib_overview}. RLLBC-Lib is available at: \href{https://github.com/Data-Science-in-Mechanical-Engineering/RLLBC}{https://github.com/Data-Science-in-Mechanical-Engineering/RLLBC}.

\section{Overarching Design Decisions}
\label{sec:general}

A core goal of RLLBC-Lib is to underscore a continuity between tabular and deep RL approaches, while providing an easily accessible implementation style.

We follow the concept of self-contained single file implementations in Python \cite{huang2022cleanrl}. These implementations are presented via Jupyter notebooks that provide a simple interface for teaching \cite{barba2019teaching} and automated grading as discussed in Section \ref{sec:bpas}. All notebooks use one common virtual environment based on pixi-env \cite{fischer2025pixi} for a simple handling of dependencies. Further, interaction between the agent and its environment is handled via the common gymnasium interface \cite{towers2026gymnasium}.

All notebooks follow a common structure, comprising the building blocks
\begin{compactitem}
    \item Agent and environment setup;
    \item Hyperparameter setting;
    \item Training loop; and
    \item Evaluation of trained agents.
\end{compactitem}
While these vary in their level of sophistication, based on the underlying algorithmic complexity, these reoccurring patterns underscore similarities between tabular and deep RL methods. For instance, parallels between tabular Q-learning \cite{watkins1992q} and deep Q-Networks \cite{Mnih2015Feb} can be drawn. The latter requires substantially more involved handling of interaction data and value approximation. Still, it inherits the greedy maximization of a target policy and an $\epsilon$-greedy exploration scheme from the former.
    
Most importantly, markdown blocks of the Jupyter notebooks are used to provide a detailed explanation of the algorithmic implementation. We reference the relevant literature and explain how certain aspects of pseudocode algorithms or relevant equations are implemented. While \cite{sutton2018reinforcement} serves as the reference for most tabular methods, we discuss the respective publications for the deep RL algorithms. This underscores similarities and differences between the respective approaches.
\section{Dynamic Programming and Tabular Reinforcement Learning
}

\begin{figure}[tb]
  \centering
    \begin{subfigure}[t]{0.4\textwidth}
    \centering
    \includegraphics[width=1\linewidth]{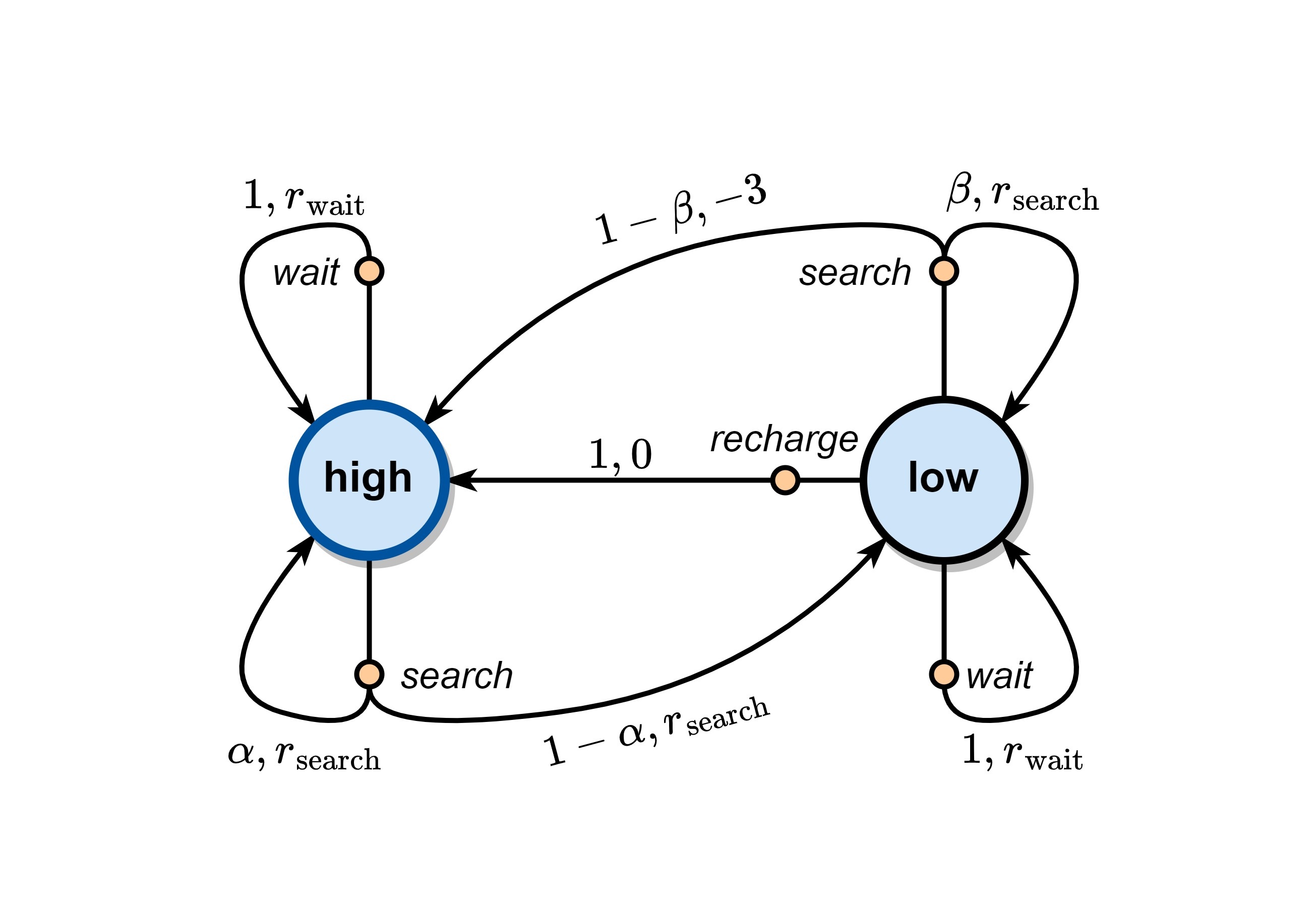}
    \subcaption{Recycling Robot}
    \label{fig:recycling_robot}
  \end{subfigure}\hfill
  \begin{subfigure}[t]{0.59\textwidth}
    \centering
    \includegraphics[width=0.42\linewidth]{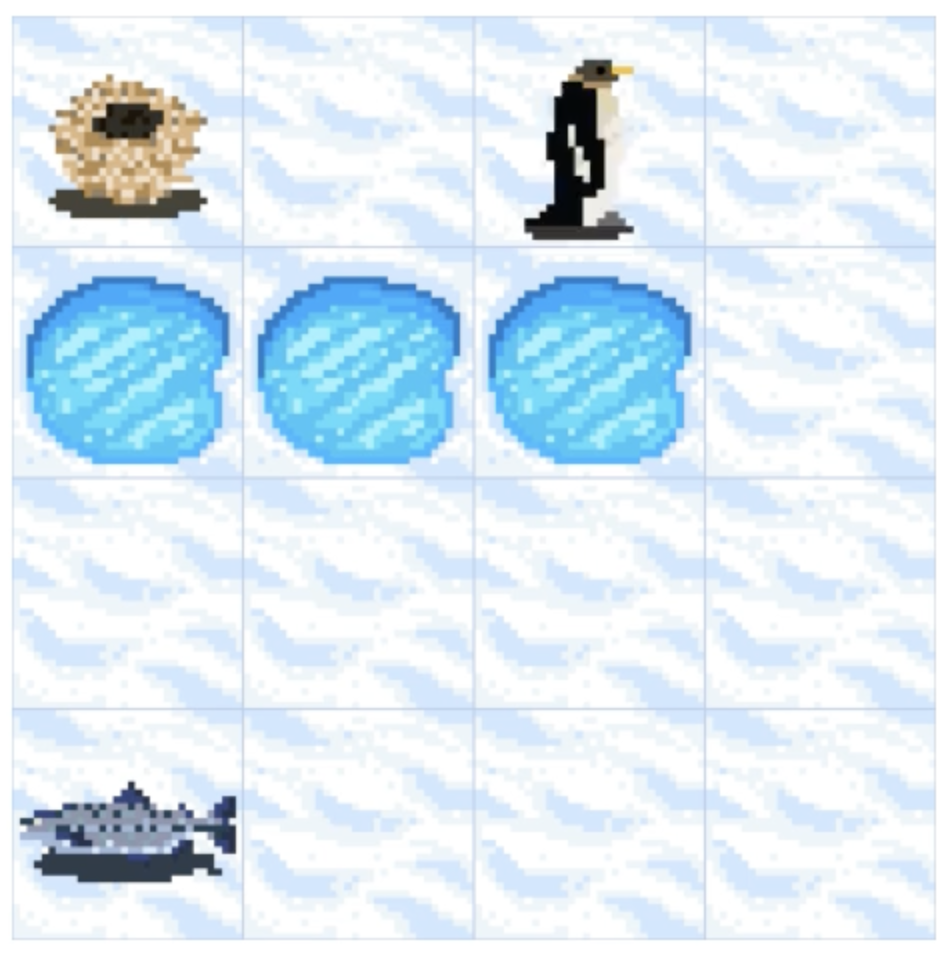}
    \includegraphics[width=0.5\linewidth]{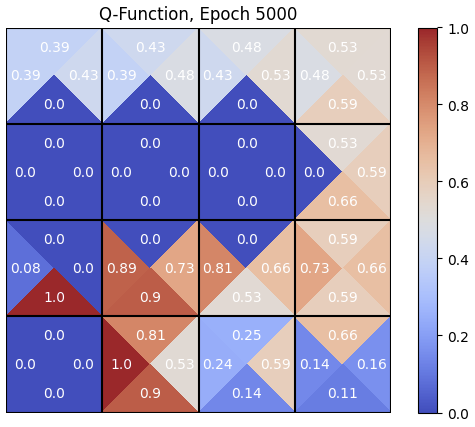}
    \subcaption{State Action Value Function Visualization}
    \label{fig:tab_q_val}
  \end{subfigure}

  \caption{Examples of Illustrative Tabular Implementations. (a) \textit{Custom implementation of the recycling robot example in \cite{sutton2018reinforcement} to explain MDP dynamics and value functions.} (b) \textit{Each state of the grid world is separated into triangles representing the action space $\mathcal{A} = \{\leftarrow, \uparrow, \rightarrow, \downarrow \}$. Plotting the state action values via color coding then illustrates learning dynamics.}}
  \label{fig:illustrations}
\end{figure}

Following the didactic concept of \cite{sutton2018reinforcement}, we first discuss tabular settings, i.e., finite Markov decision processes (MDPs) with discrete states $S$ and actions $A$. Such problems are comparably easy to handle and typically come with convergence guarantees, making them more approachable than their continuous counterparts.

We provide custom implementations of algorithms presented in \cite{sutton2018reinforcement}, introducing the fundamental concepts of dynamic programming, temporal difference learning, and Monte-Carlo methods. A comprehensive list of algorithms is provided under \texttt{tabular\_examples} in Appendix \ref{app:filetree}.

As discussed in Section \ref{sec:general}, we put emphasis on explanations and illustrations, with two examples provided in Figure \ref{fig:illustrations}. As depicted in Figure \ref{fig:recycling_robot}, we provide a gymnasium \cite{towers2026gymnasium} implementation of the recycling robot (Example 3.3 in \cite{sutton2018reinforcement}). Its inherent stochasticity and state dependent action space make it an interesting, yet low dimensional problem. We use it as a running example to illustrate the fundamentals of MDP dynamics and value function learning.

Furthermore, we provide a visualization of state action value functions for grid world environments, as illustrated in Figure \ref{fig:tab_q_val}. The state action value function estimates the expected return to go under the current policy, given a certain state action pair. Here, we visualize it for an adapted version of the \texttt{FrozenLake} environment \cite{towers2026gymnasium}. The agent is illustrated by the penguin\footnote{The depicted penguin represents Henning, the mascot of the Mechanical Engineering Student Council at RWTH Aachen University. Additional information can be found \href{https://fsmb.rwth-aachen.de/fachschaft/fachschaftsrat/}{here}.}. It starts from its nest and tries to reach the fish by navigating the frozen lake without falling into the holes. States are represented by tiles, while the action space $\mathcal{A} = \{\leftarrow, \uparrow, \rightarrow, \downarrow \}$ comprises walking in four different directions. Thus, each triangle on the right represents a particular state action pair. The corresponding state action value estimate is indicated by a rounded numerical value and a color. This allows us to plot the full state action value function at different points throughout training, which illustrates the  learning dynamics of the respective tabular algorithms.

\section{Deep Reinforcement Learning}

Deep RL notebooks follow a similar general structure as their tabular counterparts, while addressing high dimensional or continuous problems. The implementations are largely based on CleanRL \cite{huang2022cleanrl}, using PyTorch \cite{paszke2017automatic} for neural network training.
A comprehensive list of provided algorithms can be found under \texttt{deep\_examples} in Appendix \ref{app:filetree}.

\begin{figure}[tb]
  \centering
    \begin{subfigure}[t]{0.4\textwidth}
    \centering
    \includegraphics[width=0.9\linewidth]{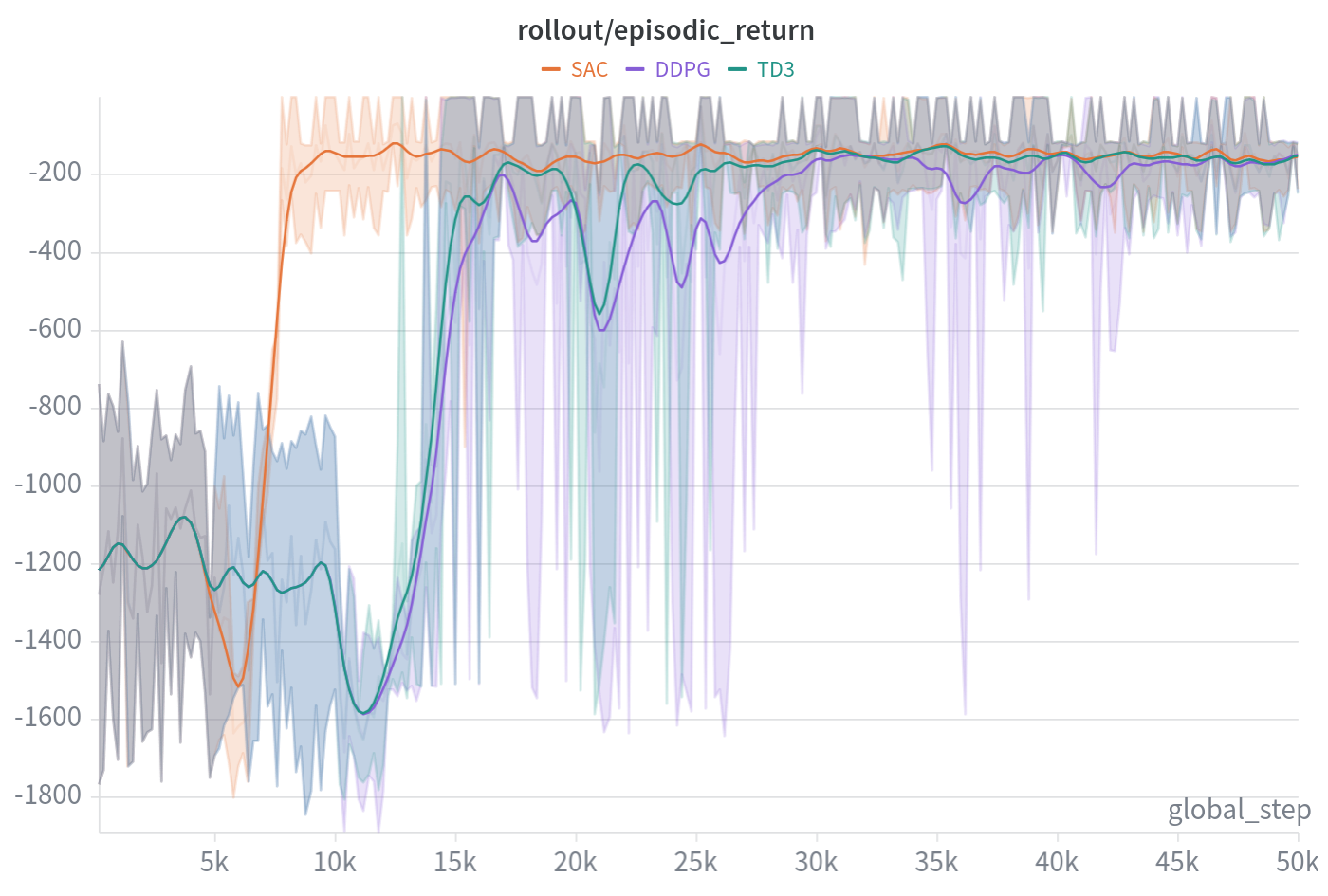}
    \subcaption{Performance Logging and Comparison}
    \label{fig:returns}
  \end{subfigure}\hfill
  \begin{subfigure}[t]{0.6\textwidth}
    \centering
    \includegraphics[width=\linewidth]{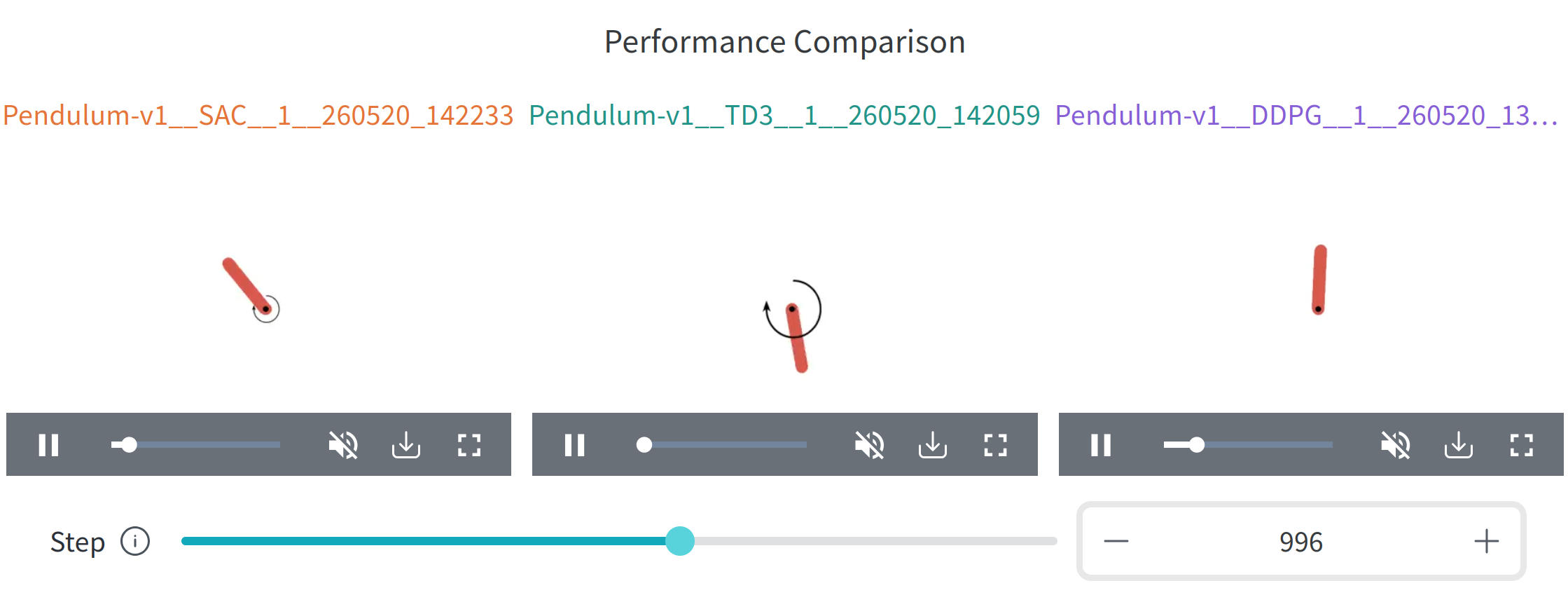}
    \subcaption{Visual Inspection of Agent Performance}
    \label{fig:video_logs}
  \end{subfigure}

  \caption{Logging of Deep RL Experiment Data. \textit{ A further objective is to introduce students to deep RL workflows. The W\&B integration allows managing large amounts of experiment data. Examples are} (a) \textit{ logging relevant quantities over several random seeds to compare algorithms or hyperparameter settings, and} (b) \textit{recording videos of agent performance to inspect failure scenarios or connect a certain return value to the corresponding behavior.}}
  \label{fig:wandb}
\end{figure}

Similar to the tabular notebooks, we provide detailed information about relevant implementation details. This emphasizes similarities and differences between value-based \cite{Mnih2015Feb}, policy-gradient \cite{Williams1992May}, as well as on- \cite{mnih2016asynchronous, pmlr-v37-schulman15, schulman2017proximal} and off-policy actor critic \cite{pmlr-v32-silver14, fujimoto2018addressing, haarnoja2018soft} approaches.

Further, we introduce students to common deep RL workflows that help monitor training and compare results. Relevant quantities such as return or temporal difference losses are logged via tensorboard \cite{abadi2015tensorflow} and optionally via W\&B \cite{biewald2020experiment}. This allows students to evaluate training performance in real time and connect lecture content with empirical observations. For instance, they can directly see how tuning trust regions influences training stability. Hierarchically organized logs allow tracking the effect of hyperparameter settings on learning performance. Further, videos of agent performance can be logged to W\&B projects and visually inspected to connect return curves to agent behavior, as depicted in Figure \ref{fig:wandb}.

Agent evaluation is handled at a configurable frequency over a variable number of episodes. This helps students understand how exploration introduces stochasticity in agent performance.
Additionally, the notebooks provide the functionality to save and load checkpoints of agents. In particular, the best performing version of the agent is saved. Consequently, students can load a trained agent after a computationally expensive training run. They can then, e.g., compare it to other agents or investigate its performance and failure scenarios leveraging the aforementioned video recording function.

\section{Additional Material}

The tabular and deep RL libraries with a consistent design concept represent the core functionality of RLLBC-Lib. Beyond that, we provide additional material for illustrating key concepts of RL and contrasting it with other LBC approaches. Further, we developed programming assignments with automated grading for scalable teaching. In the following, we provide an overview of additional materials with examples depicted in Figure \ref{fig:class_examples}.

\subsection{Class Notebooks}
\label{sec:class_ex}
\begin{figure}[tb]
  \centering
    \begin{subfigure}[t]{0.225\textwidth}
    \centering
    \includegraphics[width=1\linewidth]{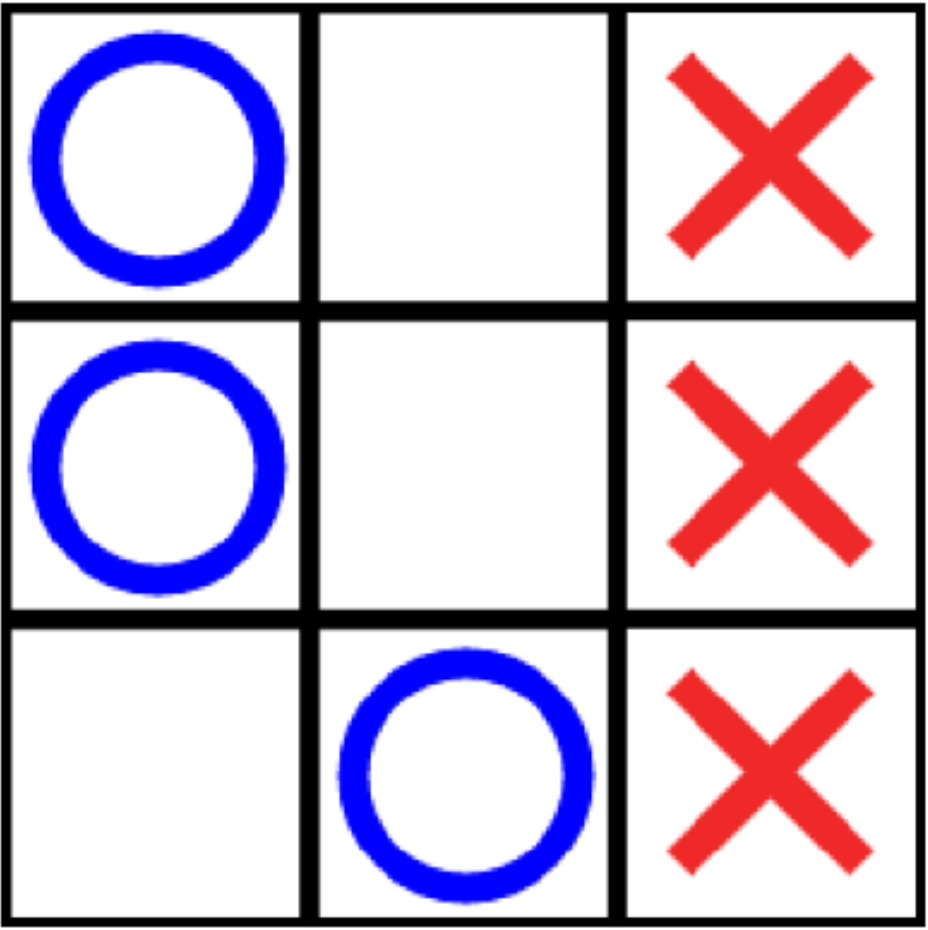}
    \subcaption{Play against Agent}
    \label{fig:tictactoe}
  \end{subfigure}\hfill
    \begin{subfigure}[t]{0.34\textwidth}
    \centering
    \includegraphics[width=1\linewidth]{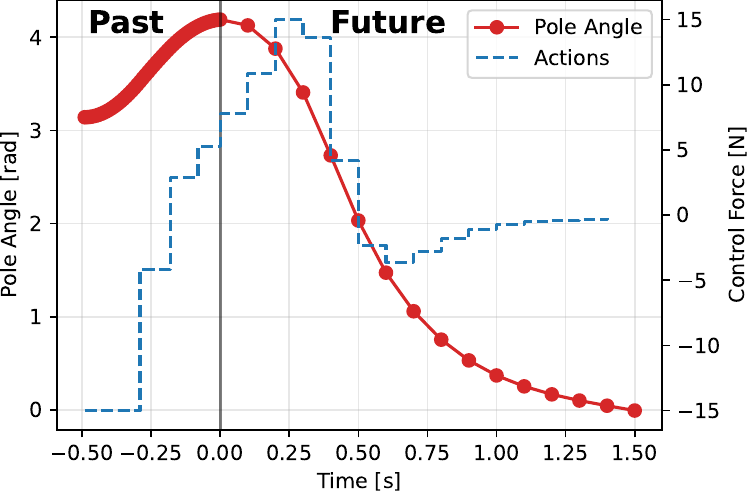
    }
    \subcaption{Model Predictive Control}
    \label{fig:lbc}
  \end{subfigure}\hfill
  \begin{subfigure}[t]{0.42\textwidth}
    \centering
    \includegraphics[width=1\linewidth]{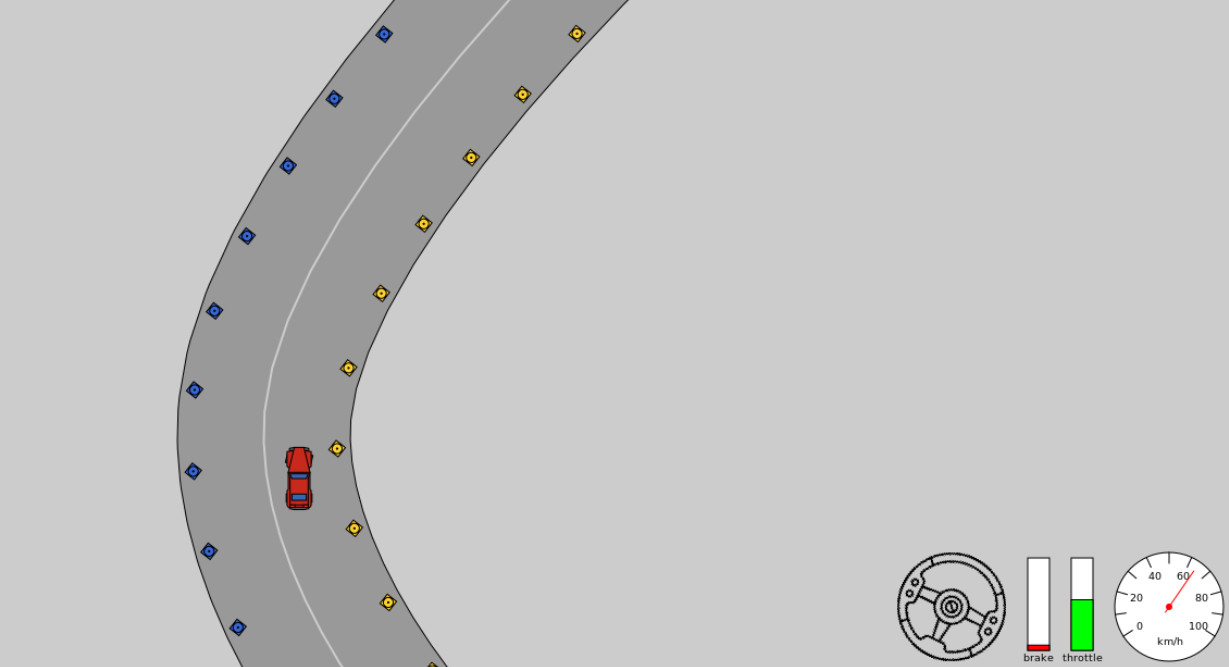}
    \subcaption{Racing Programming Assignment}
    \label{fig:racing}
  \end{subfigure}
  \caption{Additional Material. (a) \textit{Illustration examples (see Section \ref{sec:class_ex}), such as learning Tic-Tac-Toe via RL (Section 1.5 of \cite{sutton2018reinforcement})} (b) \textit{ Other learning-based control approaches (see Section \ref{sec:lbc}), such as model-predictive control.} (c) \textit{Programming assignments (see Section \ref{sec:bpas}), such as racing a car. }}
  \label{fig:class_examples}
\end{figure}

We provide illustrative examples for usage in lectures and exercise sessions that underscore certain algorithmic aspects. Many of these notebooks are implementations of examples in \cite{sutton2018reinforcement} and are either derived from the library or are standalone implementations. Figure \ref{fig:tictactoe}, for instance, depicts a notebook where an RL agent learns Tic-Tac-Toe in self play. A keyboard interface allows the lecturer play against the agent, both early in training and after convergence. Taken from Section 1.5 of \cite{sutton2018reinforcement}, this example serves to build curiosity.

\subsection{Learning-Based Control Notebooks}
\label{sec:lbc}
 As discussed above, RL falls into the greater field of LBC methods \cite{schoellig2026decade}. Control problems can be solved in a model-based or data driven way. Understanding these different approaches sharpens the intuition for when RL is well suited to a problem and when an alternative is preferable. We observe that opening the control theoretic perspective is especially helpful for computer science students that typically have a limited or no control background. Common basic knowledge also supports interdisciplinary collaboration. Beyond that, we believe transferring ideas from established fields like control theory to modern fields like deep RL can be a great source of innovation.

 We provide a brief overview of classical control topics such as state feedback control via the linear quadratic regulator~\citep{aastrom2021feedback} as well as system identification and dynamics learning~\citep{lennart1999system}.
 We further address controller tuning via Bayesian optimization~\citep{garnett2023bayesian, stenger2026decade} and model predictive control (MPC)~\citep{rawlings_model_2024}. As depicted in Figure \ref{fig:lbc}, the latter optimizes action sequences based on model-based predictions in a receding horizon fashion. Works such as \cite{Williams2017May} and \cite{Chua2018} are at the intersection of RL and MPC and underscore the fluid boundary between RL and other LBC methods.
 An overview of notebooks is provided under \texttt{lbc\_examples} in Appendix \ref{app:filetree}.

\subsection{Programming Assignments}
\label{sec:bpas}
Based on RLLBC-Lib, we developed three programming assignments for self-study.

The first assignment addresses tabular methods. From a text and graph description, students implement an MDP of a robot playing basketball as a gymnasium \cite{towers2026gymnasium} environment. They subsequently implement double Q-learning \cite{hasselt2010double} and an off-policy Monte-Carlo algorithm. The second assignment covers deep RL. Starting from Deep Q-Networks \cite{Mnih2015Feb}, students implement double DQN \cite{van2016deep} with prioritized experience replay \cite{schaul2015prioritized} and integrate generalized advantage estimation \cite{schulman2015high} into Proximal Policy Optimization \cite{schulman2017proximal}.
Both assignments mix provided code with empty cells for students to fill in. Automated grading via nbgrader \cite{ProjectJupyter2019Jan} provides a scalable solution for large courses.

The third assignment is an open racing challenge, depicted in Figure \ref{fig:racing}. It builds on an adapted version of the road environment presented in \cite{schier2023learned}. The students can pick their algorithm of choice and adapt state formulation as well as action scaling. The only decisive factor for grading the assignment is the racing performance of the submitted agent. We set up a leaderboard page, ranking the best performances of all teams to further increase motivation. This assignment closely resembles common problem settings in research and industry and gives maximum freedom to the students.

Since grading of all programming assignments is automated, we developed an upload page that continuously grades new submissions, such that students can iteratively improve their scores. For didactic reasons, we keep this page unpublished together with the assignments.
\section{Concluding Remarks}
We present RLLBC-Lib, an educational library lowering the entry barrier for students and other learners of RL. It provides a set of notebooks that guide them from the very basics of RL to state-of-the-art deep RL algorithms. The consistent notebook design between tabular and deep methods clearly underscores the parallels between the theoretical foundations and scalable approaches to RL. Extensive explanations within the notebooks make clear connections between theoretical concepts introduced in textbooks or papers and their algorithmic implementation. Usage of standard interfaces like gymnasium environments \cite{towers2026gymnasium} and W\&B logging \cite{biewald2020experiment} introduces students to best practices in RL experimentation.  Furthermore, additional notebooks illustrating core concepts of RL and contrasting them with other LBC methods sharpen the understanding of RL as a control method. Finally, RLLBC-Lib is designed such that programming assignments with automated grading can be directly derived from its implementations. This enables hands-on teaching at scale.

\fakepar{From our classroom to everyone:}
Having taught RL to several hundred students over the past years in classroom and online formats, we developed a library we wished to have had access to from the beginning. We experience RLLBC-Lib as a great aid to our teaching efforts, both in classical lectures and our massive open online courses on RL\footnote{\href{https://www.edx.org/learn/computer-science/rwth-aachen-university-reinforcement-learning-2}{https://www.edx.org/learn/computer-science/rwth-aachen-university-reinforcement-learning-2}} and LBC\footnote{\href{https://www.edx.org/learn/computer-science/rwth-aachen-university-learning-based-control}{https://www.edx.org/learn/computer-science/rwth-aachen-university-learning-based-control}}. We hope RLLBC-Lib is useful for the broader community, helping to spread the word about this exciting field.

\bibliography{rllbc}
\bibliographystyle{plain}

\newpage

\appendix
\section{Appendix}
\label{app:appendix}

\subsection{Reinforcement Learning Teaching Libraries}
\label{app:lib_overview}

\begin{table}[htbp]
\centering
\caption{A subjectively compiled non-exhaustive list of RL libraries that may be useful for teaching.}
\label{tab:rl_teaching_resources}
\small
\renewcommand{\arraystretch}{1.15}
\begin{tabular}{p{0.38\textwidth} c c c c c}
\hline
GitHub Repository
& \makecell{Jupyter\\Notebooks}
& \makecell{Dynamic\\Programming}
& \makecell{Tabular\\RL}
& \makecell{Deep\\RL}
& \makecell{Examples \& \\Explanations} \\
\hline

RLLBC-Lib (ours)
& \checkmark
& \checkmark
& \checkmark 
& \checkmark 
& \checkmark \\

EduGym
&  \checkmark
&  \checkmark
&  \checkmark
&  ( \checkmark )
&  ( \checkmark ) \\

dennybritz/\newline 
reinforcement-learning
&  \checkmark
&  \checkmark
&  \checkmark
&  ( \checkmark )
&  \\

cmendl/ \newline reinforcement-learning-course
&  \checkmark
&  \checkmark
&  \checkmark
& 
& \\

Fortuz/ \newline rl\_education
& \checkmark
& \checkmark
& \checkmark
& 
& \checkmark  \\

ShangtongZhang/ \newline reinforcement-learning-an-introduction
&  
& \checkmark 
& \checkmark 
& 
&  \\

linesd/ \newline tabular-methods
& 
&  \checkmark
&  \checkmark
&  
&  \\

prabhatnagarajan/ 
\newline table-rl
& 
&  \checkmark
&  \checkmark
&  
&  \\

lasseufpa/
\newline tabular\_rl
& 
&  \checkmark
&  \checkmark
&  
&  \\

RylinnM/
\newline Tabular-Reinforcement-Learning
& 
&  \checkmark
&  \checkmark
&  
&  \\

huggingface/ \newline deep-rl-class
& \checkmark
& 
&  ( \checkmark )
& \checkmark
& \checkmark \\

ayeenp/ \newline Deep-RL-Notebooks
& \checkmark
& 
& \checkmark
& \checkmark
& \checkmark \\

edreate/ \newline Reinforcement-Learning
&  \checkmark
&  
&  ( \checkmark )
&  ( \checkmark )
&  \\

DLR-RM/ \newline stable-baselines3
&  
&  
&  
&  \checkmark
&   \\

pytorch/ \newline rl
&  
&  
&  
&  \checkmark
&   \\

thu-ml/ \newline tianshou
& 
& 
& 
& \checkmark
&  \\

openai/ \newline spinningup
& 
& 
& 
& \checkmark
& \checkmark \\

vwxyzjn/ \newline cleanrl
&  
&  
&  
&  \checkmark
&  \checkmark \\

\hline
\end{tabular}
\end{table}

\newpage

\subsection{Library Overview}
\label{app:filetree}

\dirtree{%
.1 RLLBC/.
.2 tabular\_examples/. 
.3 dynamic\_programming/.
.4 policy\_iteration.ipynb.
.4 value\_Iteration.ipynb.
.3 tabular\_rl/.
.4 MC/.
.5 first\_visit\_mc.ipynb.
.5 every\_visit\_mc.ipynb.
.4 TD/.
.5 q-learning.ipynb.
.5 sarsa.ipynb.
.5 dyna-q.ipynb.
.2 deep\_examples/. 
.3 a2c-simple-adv.ipynb.
.3 a2c.ipynb.
.3 ddpg.ipynb.
.3 dqn.ipynb.
.3 reinforce.ipynb.
.3 sac.ipynb.
.3 td3.ipynb.
.3 trpo-simple-adv.ipynb.
.3 trpo.ipynb.
.2 lbc\_examples/.
.3 lqr.ipynb.
.3 mpc.ipynb.
.3 dynamics\_learning\_discrete.ipynb.
.3 bo.ipynb.
.2 class\_examples/. 
.3 MDP and value functions.
    .4 bellman\_opt\_eq.ipynb.
    .4 markov\_process.ipynb.
    .4 recycling\_bot\_value\_function.ipynb.
    .4 tic-tac-toe.ipynb.
.3 dynamic\_programming.
    .4 dp\_gridworld.ipynb.
    .4 dp\_gridworld2.ipynb.
.3 monte carlo methods.
    .4 BlackJack.ipynb.
.3 temporal difference learning.
    .4 cliffwalking\_on\_vs\_off-policy.ipynb.
    .4 td0\_vs\_constant\_alpha\_mc.ipynb.
    .4 td\_vs\_mc\_control.ipynb.
.3 Dyna-Q.
    .4 Dyna-Q\_vs\_Q-Learning.ipynb.
.3 function approximation.
    .4 linear\_sarsa.ipynb.
    .4 nonlinear\_approximation.ipynb.
    .4 random\_walk.ipynb.
.3 learning-based control.
    .4 Cartpole\_NARX.ipynb.
    .4 Cartpole\_lqr.ipynb.
    .4 lin\_sys\_id\_oscillator.ipynb.
}

\end{document}